\documentclass[10pt,twocolumn,letterpaper]{article}

\usepackage[pagenumbers]{wacv} 

\usepackage{booktabs}%
\usepackage{algorithm}%
\usepackage{algorithmicx}%
\usepackage{algpseudocode}%

\usepackage{amsmath}
\usepackage{booktabs}       
\usepackage{amsfonts}       
\usepackage{nicefrac}       
\usepackage{xcolor}         
\usepackage{multirow}
\usepackage{booktabs}
\usepackage{multirow}
\usepackage{graphicx}
\usepackage{wrapfig}
\usepackage[table]{xcolor}
\usepackage{bm} 
\usepackage{algorithm}
\usepackage{subcaption}
\usepackage{makecell}

\usepackage{adjustbox}
\usepackage{placeins}
\usepackage{wrapfig}
\usepackage{mathtools}
\usepackage{pifont}
\usepackage{algorithm}
\usepackage{algpseudocode}
\algtext*{EndFor}

\definecolor{ForestGreen}{RGB}{34,139,34}
\definecolor{ourgray}{gray}{0.93}
 
\newcommand{\cmark}{\textcolor{ForestGreen}{\ding{51}}}
\newcommand{\xmark}{\textcolor{black!35}{\ding{55}}}
 
\newcommand{\val}[2]{$#1$\,\textcolor{black!45}{$\pm#2$}}
\newcommand{\bval}[2]{$\mathbf{#1}$\,\textcolor{black!45}{$\pm#2$}}
\newcommand{\uval}[2]{$\underline{#1}$\,\textcolor{black!45}{$\pm#2$}}
\newcommand{\na}{\textcolor{black!40}{--}}
\newcommand{\up}{\raisebox{0.15ex}{$\uparrow$}}
\newcommand{\dn}{\raisebox{0.15ex}{$\downarrow$}}

\definecolor{wacvblue}{rgb}{0.21,0.49,0.74}
\usepackage[pagebackref,breaklinks,colorlinks,allcolors=wacvblue]{hyperref}

\def\wacvPaperID{2842} 
\def\confName{WACV}
\def\confYear{2027}

\title{UniDR: Learning To Generate Anatomical Deformations From Images
via Unified Diffusion Modeling and Registration}

\author{Swakshar Deb$^{1}$, Tonmoy Hossain$^{2}$, Tom Fletcher$^{1}$, Miaomiao Zhang$^{1}$ \\
$^{1}$Department of Electrical and Computer Engineering, University of Virginia, USA\\
$^{2}$Harvard Medical School and Massachusetts General Hospital, USA\\
{\tt\small \{swd9tc, ptf8v, mz8rr\}@virginia.edu}
}

\begin{document}
\maketitle
\begin{abstract}
This paper presents a novel framework, Unified Diffusion modeling and Registration (UniDR), which learns to generate arbitrary anatomical deformations from images with preserved topology. Existing approaches often model the generative process separately from registration-based deformation estimation; hence making them prone to propagated errors from the registration stage. To address this limitation, we propose a unified paradigm based on a conditional diffusion model over deformation fields that are jointly optimized from the registration process. A principled registration-guided score-matching objective function is then introduced to align the diffusion process with the deformation distribution induced by registration. As a result, our UniDR establishes an explicit connection between generation and registration: the diffusion model enables exploration of the space of plausible anatomical deformations, whereas registration provides structured supervision for learning transformations consistent with observed images. Once trained, UniDR is able to sample diverse, anatomically plausible transformations conditioned on any single input image. We evaluate UniDR across multiple real-world applications and demonstrate its effectiveness in generating high-fidelity image samples compared to the state-of-the-art. Additionally, we demonstrate that these generated samples improve performance in downstream medical image analysis and vision tasks. Our code is available on https://github.com/SwaksharDeb/UniDR.

\end{abstract}
    
\section{Introduction}
\label{sec:intro}

Deformation fields mathematically encode smooth spatial transformations that establish dense correspondences between images~\cite{pfeifer2006three,zhang2016statistical,thompson2002framework,grenander1998computational}. They provide a powerful and interpretable representation for modeling and quantifying geometric shape and shape variations in a wide range of image analysis tasks, including image registration~\cite{zhang2019fast,zhang2017frequency,balakrishnan2019voxelmorph}, motion tracking~\cite{liao2016temporal}, geometric data augmentation~\cite{joshi2004unbiased,hossain2025mgaug}, and cross-modal image alignment~\cite{wang2022multi}, where understanding how one image transforms into another is essential. In contrast to conventional intensity-based image representations, deformation fields explicitly characterize the underlying geometric structures of objects with desirable properties (i.e. diffeomorphisms) to ensure smooth and invertible smooth mappings to preserve object topology~\cite{beg2002computational,miller2003metric,leslie1967differential}. This makes deformation models particularly suitable for studying structural and functional changes in complex biological systems. For example, in medical imaging, deformation fields play a central role in quantifying anatomical remodeling associated with disease progression, including neurodegenerative disorders~\cite{davatzikos2001framework,fiot2014longitudinal,miller2006geodesic} and cardiac synchronization~\cite{hossain2025invariant,wiputra2020cardiac,seo2013computing}.
Furthermore, deformation fields provide a topology-aware representation of shape variation that uniquely quantifies how anatomical structures deform and evolve across subjects, time, and disease states, which has established as a cornerstone of computational anatomy~\cite{beg2002computational,niethammer2011geodesic,zhang2016low}, statistical shape analysis~\cite{zhang2016statistical,arsigny2006log,qiu2011principal,golland2001deformation}, and object tracking~\cite{zhuang2022review,pedersoli2015elastic,zhong2000object,sengupta2021visual}.

\begin{figure}[!t]
    \centering
    \includegraphics[width=\linewidth, trim={0cm 0.0cm 0.0cm 0.0cm}]{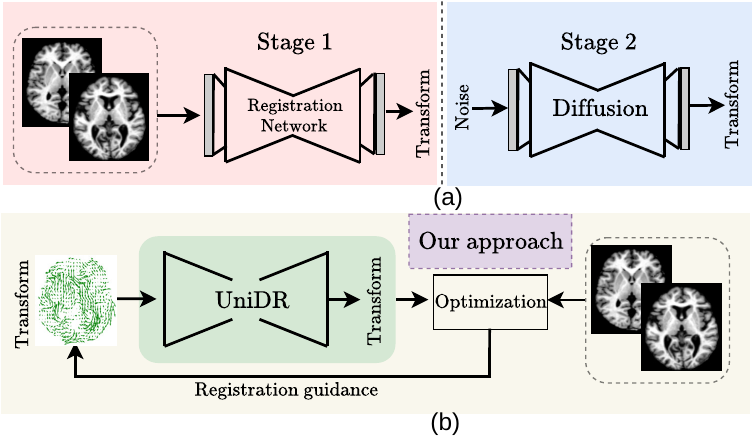}
    \caption{An overview of (a) existing two-step generative deformation model vs. (b) our proposed approach UniDR.} 
    \label{fig:intro_fig}
\end{figure}

{\em Generative deformation models} have emerged as a new line of research that aims to learn the distribution of deformation fields from images~\cite{jayakumar2023sadir,starck2025diff,jayakumar2024tpie,wu2025ldm}. These methods shift the paradigm to capture complex geometric transformations that govern how anatomical structures change with preserved topology. By explicitly modeling deformation processes, these generative deformation models for the first time provide a principled framework for representing and quantifying the geometry of synthesized images. Early efforts in this area employed variational autoencoders to learn compact representations of deformation fields by imposing Gaussian priors in the latent deformation space~\cite{dalca2019learning,hossain2025mgaug,krebs2019learning}. Later studies incorporated adversarial learning to enforce more realistic deformation priors and improve the anatomical plausibility of synthesized shapes and images~\cite{dey2021generative,bigolin2020interpretation,zheng2021symreg}. More recently, diffusion models have been introduced into deformation-space learning, achieving state-of-the-art performance in synthesizing realistic images with intact geometry~\cite{jayakumar2024tpie,wu2025ldm}, and reconstructing a complete 3D anatomy from a set of sparse input images~\cite{chen2024deformable,jayakumar2023sadir}. Despite these advances, current methods fundamentally decouple deformation estimation from generative learning. More specifically, all deformation fields are first predicted using a separate image registration network and then treated as fixed observations for training the generative model (as shown in Fig.~\ref{fig:intro_fig}). This can directly propagate registration errors into the diffusion model and potentially bias the learned data distribution~\cite{wang2025generating,jayakumar2024tpie}.


To mitigate this issue, this paper presents a new approach, UniDR, to learn the distribution of unknown deformation fields in a unified modeling of generative diffusion process with registration. With the difference from previous approaches~\cite{jayakumar2024tpie,wang2025generating,krebs2019learning} that perform diffusion in the intensity space of pairwise images, our proposed UniDR embeds the entire diffusion process in the deformation space jointly learned from the input images. {\em Once trained, UniDR can generate anatomically plausible and structurally consistent deformations conditioned on a given source image, without requiring instance-specific registration optimization used in previous methods~\cite{zhuo2024diffusereg,kim2022diffusemorph} during inference.} The main contributions of UniDR are summarized as follows:
\begin{enumerate}[(i)]
\item Develop a novel unified framework that jointly optimizes generative modeling and deformation learning, allowing bidirectional optimization during training. 
\item Introduce a principled objective function to bridge diffusion-based generative learning and deformation estimation, ensuring the learned distribution faithfully captures the underlying deformations encoded in the input images.
\item Demonstrate the effectiveness of UniDR in real-world medical applications, consistently improving the performance of downstream image analysis tasks. 
\end{enumerate}
Experimental results show that our proposed model substantially outperforms state-of-the-art generative models~\cite{wang2025generating,zhang2023adding,jayakumar2024tpie} in preserving anatomical fidelity in both brain and cardiac MRI datasets. The anatomically meaningful image generated by UniDR lead to improved performance in downstream tasks, highlighting its potential impact in real-world applications.

\section{Background: Deriving Deformations From Pairwise Images}
Let $\Omega = \mathbb{R}^d / \mathbb{Z}^d$ be a $d$-dimensional torus domain with periodic boundary conditions. Consider any two images $S$ and $T$ defined on the torus domain $\Omega$ ($S(x), T(x): x \in \Omega \rightarrow \mathbb{R}$), where $S$ denotes the source (moving) image and $T$ denotes the target (fixed) image. The goal of deformable image registration~\cite{beg2005computing,arsigny2006log,avants2008symmetric} is to find a deformable transformation, $\psi(x): \Omega \rightarrow \Omega$, such that the deformed source image, $S \circ \psi^{-1}$, aligns with $T$, where $\circ$ denotes the interpolation of $S$ at transformed coordinates. The deformation $\psi$ can be broadly categorized into two types: (i) global transformations such as affine mappings (e.g., rotation, translation, scaling), and (ii) pixel-wise deformations that captures complex local anatomical variability. Both can be formulated by Lie groups and their associated Lie algebras, which provide a principled mathematical foundation for modeling smooth geometric transformations~\cite{miller2003metric,leslie1967differential,rosenfeld2013geometry}.

A set of invertible {\bf affine transformations} of $\mathbb{R}^d$ forms a Lie group, denoted by $\mathrm{Aff}(\mathbb{R}^d)$. Specifically, an affine transformation $\psi_a(x)$ can be formulated as
\begin{equation}\label{eq:affine}
\small
    \psi_a(x) = A x + b, \quad A \in \mathrm{GL}(d),\; b \in \mathbb{R}^d,
\end{equation}
where $\mathrm{GL}(d)$ is the general linear group of invertible $d \times d$ matrices, and $b$ is a translation vector. This group is a matrix Lie group, and its associated Lie algebra, denoted $\mathfrak{aff}(\mathbb{R}^d)$, consists of all matrices that represent the infinitesimal generators of affine transformations~\cite{rosenfeld2013geometry,freudenthal1964lie}. \\
\noindent {\bf Deformable transformations} are often modeled as diffeomorphisms (denoted as $\psi_{\text{diff}}(x)$), which forms an infinite-dimensional Lie group $\mathrm{Diff}(\Omega)$ - a group of smooth and invertible mappings with smooth inverses. The associated Lie algebra forms a tangent space $V = T_{id} \mathrm{Diff}(\Omega)$ at the identity $id \in \mathrm{Diff}(\Omega)$. Such a tangent space, composed of smooth velocity fields, $v \in V$, provides a compact, continuous, and mathematically well-structured linear representation of complex deformations~\cite{arsigny2006log,beg2005computing}. It is widely used as an efficient parameterization of deformation fields under a large diffeomorphic deformation metric mapping~\cite{beg2005computing}, or a stationary velocity field (SVF) that remains constant over time~\cite{arsigny2006log}. While we employ SVF in this paper, our framework is easily applicable to the other.   

For a stationary velocity field $v$, the diffeomorphisms $\{\psi_{\text{diff}}(\tau,x)\}, \tau \in [0, 1]$, are generated as solutions to the ordinary differential equation:
\begin{equation}
\label{eq:svf}
\frac{d\psi_{\text{diff}}(\tau,x)}{d\tau} = v \circ \psi_{\text{diff}}(\tau,x), \quad \text{s.t.} \quad \psi_{\text{diff}}(0,x) = x.
\end{equation}
The solution of Eq.~\eqref{eq:svf} is identified as an exponential map, i.e., $\psi_{\text{diff}}(\tau,x) = \exp(v,\tau)$. The final deformation $\psi_{\text{diff}}(1,x)$ can be computed through a recursive composition using a scaling and squaring scheme~\cite{arsigny2006log}. For notational simplicity, we denote $\psi_{\text{diff}}(1,x) \triangleq \psi_{\text{diff}}$ by omitting the index of time and spatial coordinates in following sections. 

Finally, a registration loss is typically used to estimate velocity fields from given pairwise images as follows
\begin{equation}
\label{eq:energy}
\mathcal{L}_{\text{reg}}(v) = \text{Dist}(S \circ \psi_{\text{diff}}^{-1}(v), T) + \gamma_w \, \text{Reg}(v), \text{s.t. Eq.~\eqref{eq:svf}},
\end{equation}
where $\text{Reg}(\cdot) = \| \nabla v \|$ serves as a regularization term, weighted by the parameter $\gamma_w$, to ensure smoothness of the transformation fields. The $\text{Dist}(\cdot, \cdot)$ is a distance function that measures the dissimilarity between images, e.g., sum-of-squared differences~\cite{beg2005computing}, normalized cross correlation~\cite{avants2008symmetric}, and mutual information~\cite{wells1996multi}.

\section{Our Method: UniDR}
\begin{figure*}[!htbp]
    \centering
    \includegraphics[width=0.8\linewidth, trim=0 0 0 0, clip]{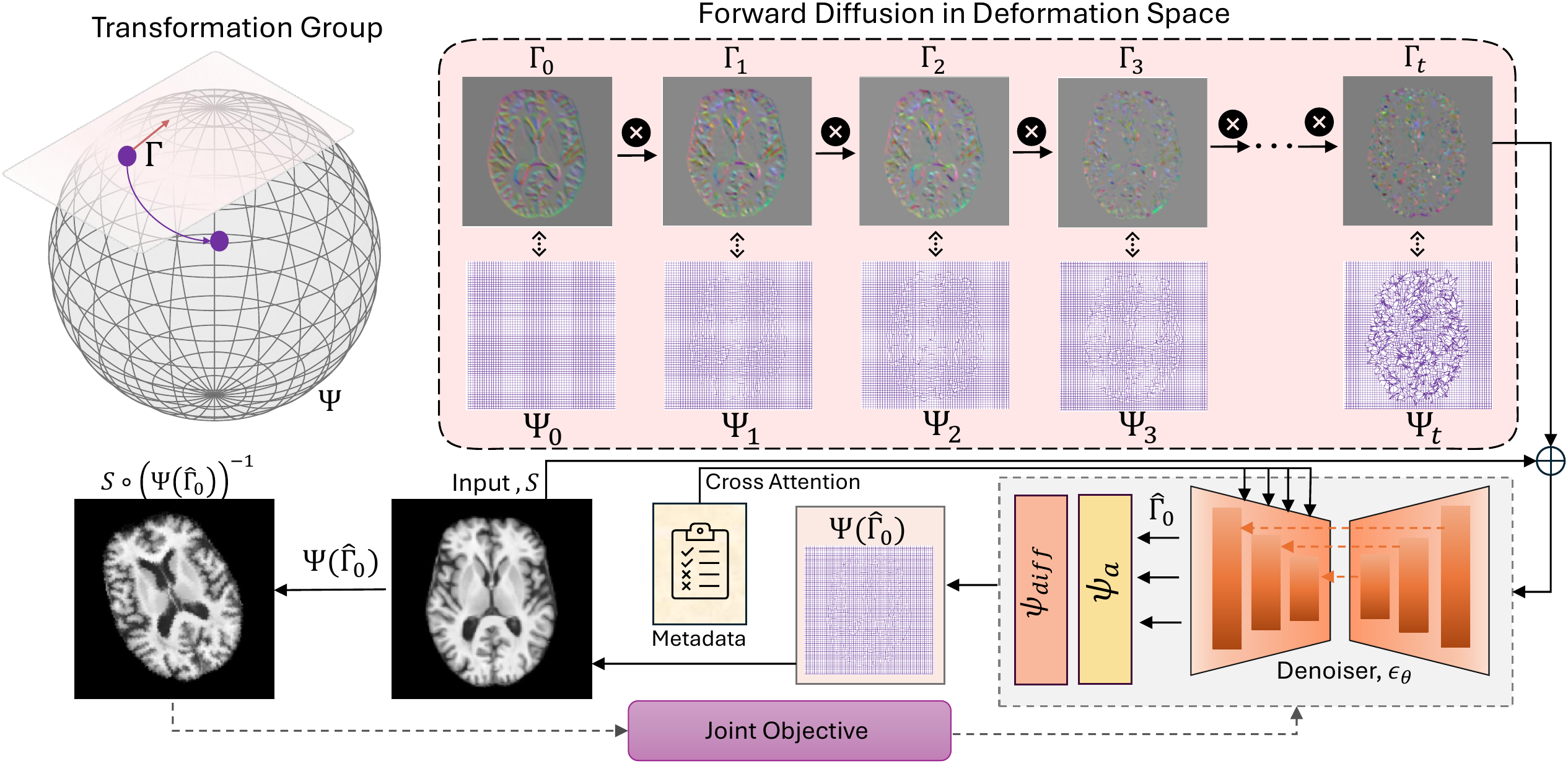}
    \caption{An overview of our proposed network architecture of UniDR. $\oplus$ indicates the channel-wise concatenation operation.}
    \label{fig:model}
\end{figure*}

This section proposes a novel generation model, UniDR, that for the first time unifies the generation process of deformations with registration schemes. Our model has two primary components: (i) a forward diffusion in the parameterized deformation space (featuring both global and local transformations) in a single step; and (ii) a reverse process that learns to generate target transformations guided by registration loss. Once trained, UniDR is able to generate all plausible and anatomically consistent deformations conditioned on the information of the input source image without the need of registration. An overview of our proposed network architecture is shown in Fig.~\ref{fig:model}.

\noindent{\bf Problem Setup.} Given a set of $N$ pairwise images $\{S^i,T^i\}_{i=1}^{N}$, we model the transformation between each source image $S^i$ and target image $T^i$ using a latent velocity field $v^i$ and a latent affine transformation parameter $A^i$. The velocity field $v^i$ induces a diffeomorphic transformation $\psi_{\mathrm{diff}}^i$ through Eq.~\eqref{eq:svf}, while $A^i$ parameterizes an affine transformation $\psi_a^i$ through Eq.~\eqref{eq:affine}. Following the group composition property of diffeomorphisms and affine transformations~\cite{rosenfeld2013geometry}, we define the overall transformation as a composite transformation $\Psi^i = \psi_a^i \circ \psi_{\mathrm{diff}}^i$. The target image is assumed to be generated by deforming the source image through the composite transformation $\Psi^i$. This yields the conditional likelihood as $p(T^i \mid S^i,\Gamma^i)$, which measures the consistency between the target image and the transformed source image $S^i \circ (\Psi^i)^{-1}$. 

Defining the latent transformation parameters as $\Gamma^i \triangleq \{v^i, A^i\}$, we then introduce a conditional diffusion prior $p_\theta(\Gamma^i | S^i)$, parameterized by $\theta$, to model the distribution of anatomically plausible transformations conditioned on structural information of the source image. 

 By the Baye's rule, we obtain a posterior inference of the latent deformation parameter formulated as
\begin{equation}
\small
\prod_{i=1}^N p(\Gamma^i \mid S^i,T^i) \propto \prod_{i=1}^N p(T^i \mid S^i,\Gamma^i) \, p_\theta(\Gamma^i \mid S^i). 
\label{eq:uniRDpost}
\end{equation}
This provides a unified generative modeling paradigm that captures the conditional distribution of anatomically plausible transformations, while leveraging the registration likelihood to guide learning toward the underlying deformations between images.

Taking the logarithm of Eq.~\eqref{eq:uniRDpost}, we arrive at the learning objective underlying our registration-guided score-matching formulation as
\begin{equation}
\label{eq:logUniDR}
\small
\sum_{i=1}^N \log p(\Gamma^i \mid S^i,T^i) \approx \sum_{i=1}^N \log p(T^i \mid S^i,\Gamma^i) + \log p_\theta(\Gamma^i \mid S^i). 
\end{equation}

\subsection{Registration-guided Diffusion Inference} 
\label{sec:forward}
To maximize the log posterior in Eq.~\eqref{eq:logUniDR}, we introduce a score-based diffusion process over the latent transformation parameters to enable tractable inference. Specifically, a {\em forward diffusion process} gradually perturbs an initial latent transformation, $\Gamma$, and produces a sequence of increasingly noisy variables. The {\em reverse process} then learns to recover the underlying transformation by maximizing the posterior score. Since the optimal transformation is unknown, our score-based denoising process is jointly optimized with the registration objective during training. This allows the recovered transformation to be progressively refined toward a solution that maximizes the log posterior. To simplify the notation, we omit the subject index $i$ in following sections.
\subsubsection{Forward Diffusion in Deformation Space} 
Similar to \cite{ho2020denoising,song2021denoising}, at each diffusion step $t \in \{1, \cdots ,\tau\}$, our forward process defines a Markov chain that progressively perturbs the deformable and affine transformation parameters, $\Gamma_t$, through noise injection, $\epsilon_{\Gamma} \sim \mathcal{N}(0, \mathbf{I})$. Note that we perform the stochastic perturbation directly in the affine parameter space rather than on the affine transformation matrix. This preserves the interpretability of individual transformation components (e.g., rotation angles, translation, and scaling factors) throughout the diffusion process.

Given $\Gamma_0$, we can formulate the forward diffusion process as a single-step computation as
\begin{equation}
   \Gamma_{t} \leftarrow \sqrt{\bar{\alpha}_t} \, \Gamma_{0} + \sqrt{1 - \bar{\alpha}_t} \, \epsilon_{\Gamma},
   \label{eq:forward}
\end{equation}
where $\bar{\alpha}_t=\prod_{s=1}^{t}(1-\beta_s)$ controls the signal-to-noise ratio and $\beta_s \in (0,1)$ is the predefined noise scheduler that is gradually increasing over each diffusion step.


\subsubsection{Registration-guided Reverse Process}
Following a similar principle of~\cite{anderson1982reverse,song2021scorebased}, we are now ready to formulate posterior inference through a registration-guided reverse-time stochastic differential equation (SDE). Specifically, the reverse process is governed by the posterior score defined in Eq.~\eqref{eq:logUniDR}. Substituting this posterior score into the discretized reverse-time SDE gives us
\begin{equation}
\small
  \Delta \Gamma_{t} = \Big[ -\frac{1}{2}\beta_t \Gamma_t - \beta_t \nabla_{\Gamma} \log p_t\! (\Gamma \mid {S}, T ) \Big] \Delta t + \sqrt{\beta_t} \mathbf{z}
\label{eq:sde}
\end{equation}
where $-\frac{1}{2} \beta_t \Gamma_t$ represents a drift coefficient, $\sqrt{\beta_t}$ denotes the diffusion coefficient. Both $\Delta \Gamma_t$ and $\Delta t$ are discretized backward in time and $\mathbf{z} \sim \mathcal{N}({0, \mathbf{I}})$.

Combining Eq.~\eqref{eq:logUniDR} and Eq.~\eqref{eq:sde}, we can derive a joint learning objective as a minimization of registration-guided score-matching function as
\begin{align}
\small
\mathcal{L}(\Gamma, \theta) &= - \sum_{i=1}^N \log p(T \mid S,\Gamma) + \log p_\theta(\Gamma \mid S) \nonumber\\
                            &\propto \!-\! \sum_{i=1}^N [ \log \, 
       \exp \left(-\tfrac{1}{\gamma_a} (\mathcal{L}_{\text{reg}}(\Gamma) \right) \!
+ \log p_\theta(\Gamma \mid S)] \nonumber\\      
                            & \simeq \, \frac{1}{\gamma_a} \mathbb{E}\big[\| S \circ(\Psi(\Gamma))^{-1} - T \|^2_2  + \gamma_w \|\nabla v \|_2^2 \big] \nonumber \\
                            &\quad + \mathbb{E}\big[\|\epsilon_\Gamma -\epsilon_\theta \|_2^2\big],
\label{eq:loss}
\end{align}
where $\epsilon_\theta$ is a UNet-based denoising network~\cite{ho2020denoising} that estimates the injected noise $\epsilon_\Gamma$ during the forward process. \\

\noindent {\bf Optional Auxiliary Condition.} Our model UniDR can optionally incorporate auxiliary conditioning (e.g., meta data such as age, clinical variables) if available. Given an embedded conditioning representation $c$, UniDR can flexibly incorporate the auxiliary information through commonly used cross-attention mechanism~\cite{rombach2022high,jayakumar2024tpie,puglisi2024enhancing}.

\subsubsection{Optimization} 
Rather than solving the objective Eq.~\eqref{eq:loss} through instance-specific iterative optimization of both $\Gamma$ and $\theta$, we adopt an amortized inference strategy~\cite{song2021denoising,ho2020denoising,amos2023tutorial}. Specifically, we learn a shared reverse diffusion process that maps noisy transformations $\Gamma_t$ to estimates of the underlying clean transformation $\hat{\Gamma}_0$. During training, the reverse process is optimized jointly with the registration objective, such that the learned score function captures the posterior distribution of transformations that satisfy the registration objective. In this way, $\Gamma$ is no longer treated as an explicit optimization variable, but is implicitly represented through the denoising network in the conditional diffusion process. Similar to~\cite{song2021denoising}, we obtain an amortized estimator of the posterior mean of the latent transformation as
\begin{equation}
\hat{\Gamma}_0
=
\frac{1}{\sqrt{\bar{\alpha}_t}}
\left(
\Gamma_t
-
\sqrt{1-\bar{\alpha}_t}\, {\epsilon}_\theta(\Gamma_t, S, c, t)
\right).
\label{eq:reconstruct}
\end{equation}
The original joint objective in Eq.~\eqref{eq:loss} can be formulated as
\begin{equation}
\small
\min_{\theta}
\;
\mathbb{E}_{\hat{\Gamma}_0,t,\epsilon_\Gamma}
\Big[
\frac{1}{\gamma_a} \, \mathcal{L}_{\mathrm{reg}}\big(S,T;\hat{\Gamma}_0\big) +
\|\epsilon_{\Gamma} - {\epsilon}_\theta(\Gamma_t,S, c, t)\|_2^2
\Big].
\label{eq:final_objective}
\end{equation}
During training, the registration term guides the denoiser $\epsilon_{\theta}$ toward transformations consistent with the observed image pair, while the denoiser provides an amortized parameterization of the score of the prior distribution $p(\Gamma \! \mid \! S, c)$. 


\noindent {\bf Discretized Reverse Sampler.} Following the principles of~\cite{song2021scorebased}, we employ the denoiser to sample from the learned prior distribution through the discrete-time update rule
\begin{equation}
\small
  \Gamma_{t-1} = \Gamma_t + \tfrac12\beta_{t}\,\Gamma_t
  - \frac{\beta_{t}}{\sqrt{1-\bar\alpha_t}}\,{\epsilon}_\theta(\Gamma_t,S, c, t)
  + \sqrt{\beta_{t}} \, \mathbf{z}.
  \label{eq:revfinal}
\end{equation}
The overall training and sampling processes are summarized in Alg.~\ref{alg:uniDR}. 
\begin{algorithm}
\caption{Training and sampling procedure of UniDR}
\label{alg:uniDR}
\begin{algorithmic}[1]
\Statex \hspace{-1.6em}\textbf{ ** \em Training (joint optimization) **}
\State \textbf{Input:} Given $\mathcal{D}=\{S^i, T^i\}_{i=1}^N$; ${\epsilon}_\theta$; initialize $\Gamma^{(0)}$;
\For{each training step $r=0,\dots,R$}
    \State Perform forward diffusion using $\Gamma^{(r)}$ via Eq.~\eqref{eq:forward};
    \State Optimize UniDR objective  via Eq.~\eqref{eq:final_objective} to update $\theta$;
    \State $\Gamma^{(r+1)} \leftarrow\ \hat{\Gamma}_0$ via Eq. \eqref{eq:reconstruct}

\EndFor
\State \Return $\hat{\epsilon}_\theta$
\Statex \hspace{-1.6em}\textbf{\em ** Sampling **}
\State \textbf{Input:} Image $S$, trained denoiser ${\epsilon}_\theta$
\State $\Gamma_T \leftarrow \mathbf{z}; \, \, \mathbf{z} \sim\mathcal{N}(0,\mathbf{I})$
\For{$t=T,\dots,1$}
    \State $\mathbf{z} \sim\mathcal{N}(0,\mathbf{I})$ if $t>1$ else $0$; sample $\Gamma_{t-1}$ via \eqref{eq:revfinal}
\EndFor
\State $\psi_{diff} \leftarrow\exp(v_0)$ \textcolor{lightgray}{(scaling \& squaring)}; $\Psi = \psi_a \circ \psi_{\mathrm{diff}}$;
\State $\hat T \leftarrow S \circ \Psi^{-1}$ 
\State \Return $\hat T,\Psi$
\end{algorithmic}
\end{algorithm}

\section{Experimental Evaluation}
We validate UniDR on both synthetic data (Rotated-Translated MNIST) and real-world medical imaging datasets, including 2D cardiac MRI and 3D brain MRI. We benchmark its performance primarily against state-of-the-art generative deformation methods, including MorphLDM~\cite{wang2025generating} and TPIE~\cite{jayakumar2024tpie}, which represent the current two-stage paradigm for generative deformation modeling and serve as important baselines for assessing the benefits of our proposed joint learning approach. We further compare UniDR with generative diffusion models that operate directly in the image-intensity space, including medical imaging-specific methods such as BrLP~\cite{puglisi2024enhancing} and SADM~\cite{yoon2023sadm}, as well as general-purpose diffusion frameworks such as LDM~\cite{rombach2022high} and ControlNet~\cite{zhang2023adding}. For a fair comparison, all methods are conditioned on the input image and associated metadata if necessary.
\subsection{Datasets.} 
{\bf Synthetic MNIST.} We use $5000$ handwritten digits (10 classes, $0$-$9$) from the MNIST repository~\cite{lecun1998mnist}, creating a Rotated MNIST dataset by rotating each digit between $0^\circ$-$180^\circ$ and translating by $0$-$4$ pixels. Each class includes approximately $500$ images of size $64^2$ with normalized in intensity (more details are provided in the supplementary.) \\
\noindent \textbf{3D Brain MRIs.} We use T1-weighted  MRIs of $1021$ participants with at least $4$ longitudinal visits from the ADNI repository \cite{mueller2005alzheimer}. All scans underwent skull-stripping, intensity normalization, affine registration to a common template space~\cite{brett2002problem}, and resizing to $128^3$. We consider scans from participants aged $55–92$, spanning three diagnostic classes: Cognitively Normal (CN), Alzheimer's Disease (AD), and Mild Cognitive Impairment (MCI). We utilize age, gender, and diagnosis at each visit as metadata information. To measure anatomical-specific regions that were established as biomarkers for brain diseases, we obtain ground truth segmentation masks using the FreeSurfer software~\cite{fischl2012freesurfer} that underwent expert quality control. \underline{For downstream task}, we use $414$ publicly available T1-weighted longitudinal brain MRIs with $15$ manually delineated anatomical structures from the OASIS dataset~\cite{marcus2007open}. All MRIs were resampled to the resolution of $128^3$, and followed a preprocessing pipeline similar to that used for ADNI.  \\
\noindent \textbf{2D Cardiac MRIs.} We use publicly available cine CMR images from the ACDC dataset~\cite{bernard2018deep}, which provides expert-labeled annotations of the left ventricle (LV), right ventricle, and myocardium. Disease diagnosis labels, ranging from NOR (normal) to DCM (dilated cardiomyopathy), are used as metadata. All images were resampled to an isotropic resolution of $1\text{mm}^2$, cropped to $128^2$, and temporally and spatially aligned. We use the End-Diastolic (ED) and End-Systolic (ES) frames in our experiments, resulting in a total of $1192$ images. \underline{For downstream task}, we use the LV myocardial mask of the ED and ES frames. We hold out $300$ images for the downstream segmentation task, while remaining $792$ images used for UniDR, with 554 images for training, 158 for testing and 80 for validation.


\subsection{Experiments} 
We first evaluate UniDR on the synthetic MNIST dataset, where we show the generated images with various geometric variations that include both affine and deformable transformations. Note that we exclude BrLP~\cite{puglisi2024enhancing} and SADM~\cite{yoon2023sadm} here, as these methods are specifically designed for medical imaging data.

We then evaluate our model on real-world datasets to assess its ability to generate anatomically plausible images under clinically meaningful conditions. For the longitudinal ADNI brain dataset, we use gender, disease diagnosis, and age at two longitudinal time points as metadata conditions, and evaluate whether the model can generate follow-up scans that closely match the corresponding ground-truth. For the cardiac dataset, we use the diagnosis label as the metadata condition and task the model with generating the ES frame from a given ED frame within each subject. This provides a controlled evaluation of UniDR's ability, and enable both qualitative and quantitative assessment of the similarity between the generated and ground-truth target. \\
\noindent {\bf Evaluation Metrics.} We employ evaluation metrics that assess both the quality of the generated images and the deformation fields. For image fidelity, we report the Fréchet Inception Distance (FID)~\cite{heusel2017gans} and Kernel Inception Distance (KID)~\cite{DBLP:conf/iclr/BinkowskiSAG18}, which quantify the similarity between the distributions of real and synthesized images in a learned feature space. For the synthetic MNIST, we compute these metrics using a pretrained ResNet-50 model on ImageNet~\cite{kang2023studiogan}. For the medical imaging dataset, we employ domain-specific feature extractors, using pretrained Med3D~\cite{chen2019med3d} for 3D brain MRI and RadImageNet~\cite{woodland2024feature}  for 2D cardiac MRI. To assess sample diversity, we report the Multi-Scale Structural Similarity Index Measure (MS SSIM)~\cite{friedrich2024wdm,wang2025generating} and Learned Perceptual Image Patch Similarity (LPIPS)~\cite{yang2019diversity,zhang2018unreasonable}.

Importantly, we also evaluate the Jacobian determinant (DetJac) of the generated deformation fields to directly assess whether the transformations preserve anatomical topology. A DetJac value of $1$ indicates local volume preservation, values below $1$ indicate local contraction, and values above $1$ indicate local expansion. Negative values indicate a violation of anatomical topology, corresponding to local folding or non-invertible transformations. \\
\noindent \textbf{Downstream Task.} To assess the practical utility of the generated samples, we evaluate them as data augmentation for downstream segmentation tasks on both brain and cardiac imaging datasets. Specifically, we train two representative segmentation architectures, UNet~\cite{ronneberger2015u} and Swin-UNETR~\cite{hatamizadeh2021swin}, using (i) only the available ground-truth training data and (ii) a combination of ground-truth and synthetically generated samples. Note that, since our model directly samples deformations, the same transformation can be applied to both the image and its corresponding manual segmentation mask. This naturally enables anatomically consistent, label-preserving data augmentation. We follow the 70/20/10 train/test/validation split. To evaluate downstream segmentation performance, we report the Dice Similarity Coefficient (DSC)~\cite{zou2004statistical} to assess the overlap between predicted segmentations and ground-truth annotations.  \\
\noindent \textbf{Implementation Details.} We set $\gamma_a$ and $\gamma_w$ as $1$ and $0.5$ for brain and synthetic MNIST and $0.1$ and $0.05$ for cardiac. The model is optimized using the Adam optimizer with a learning rate of $10^{-4}$, a batch size of $4$, and is trained for $200$K steps. We use a 3D UNet-based denoiser following the OpenAI/guided-diffusion repository for 3D brain MRI experiments \cite{dhariwal2021diffusion}. For the forward process, we adopt the sigmoid-beta noise scheduler. \cite{jabri2023scalable}, and for sampling, we use $1000$ denoising steps as \cite{song2021denoising}. We use 70/20/10 train/test/validation subject-wise split in all our experiments. All reported performance is averaged over $10$ independent sampling runs. See Supplementary Sec. 1 for further details regarding the evaluation metrics. 



\subsection{Overall Results}
\begin{figure}[!b]
\centering

\setlength{\tabcolsep}{5pt}
\renewcommand{\arraystretch}{1.2}
\captionof{table}{\textbf{Quantitative evaluation on Roto-Translated MNIST.}
\textbf{Transf.} indicates if a method explicitly generates transformations.}
\label{tab:results_mnist}

\resizebox{\linewidth}{!}{%
\begin{tabular}{@{}l c cccc c@{}}
\toprule
\textbf{Method} & \textbf{Transf.}
& \textbf{FID}~\dn & \textbf{KID}~\dn & \textbf{MS-SSIM}~\dn & \textbf{LPIPS}~\up
& \textbf{DetJac$_{-}$ (\%)}~\dn \\
\midrule
LDM~\cite{rombach2022high}          & \xmark & 22.110 & 0.4720 & 0.404 & 0.170 & \na \\
ControlNet~\cite{zhang2023adding}   & \xmark & 21.130 & 0.3820 & 0.417 & 0.182 & \na \\
MorphLDM~\cite{wang2025generating}  & \cmark & 18.860 & 0.0815 & 0.304 & 0.208 & 0.0930 \\
TPIE~\cite{jayakumar2024tpie}       & \cmark & \underline{18.289} & \underline{0.0801} & \underline{0.271} & \underline{0.225} & \underline{0.0780} \\
\rowcolor{ourgray}
\textbf{UniDR (Ours)}               & \cmark & \textbf{17.243} & \textbf{0.0782} & \textbf{0.239} & \textbf{0.244} & \textbf{0.0340} \\
\bottomrule
\end{tabular}}

\vspace{5pt}

\includegraphics[
    width=\linewidth,
    trim={0.0cm 0.0cm 0.0cm 0.0cm},
    clip
]{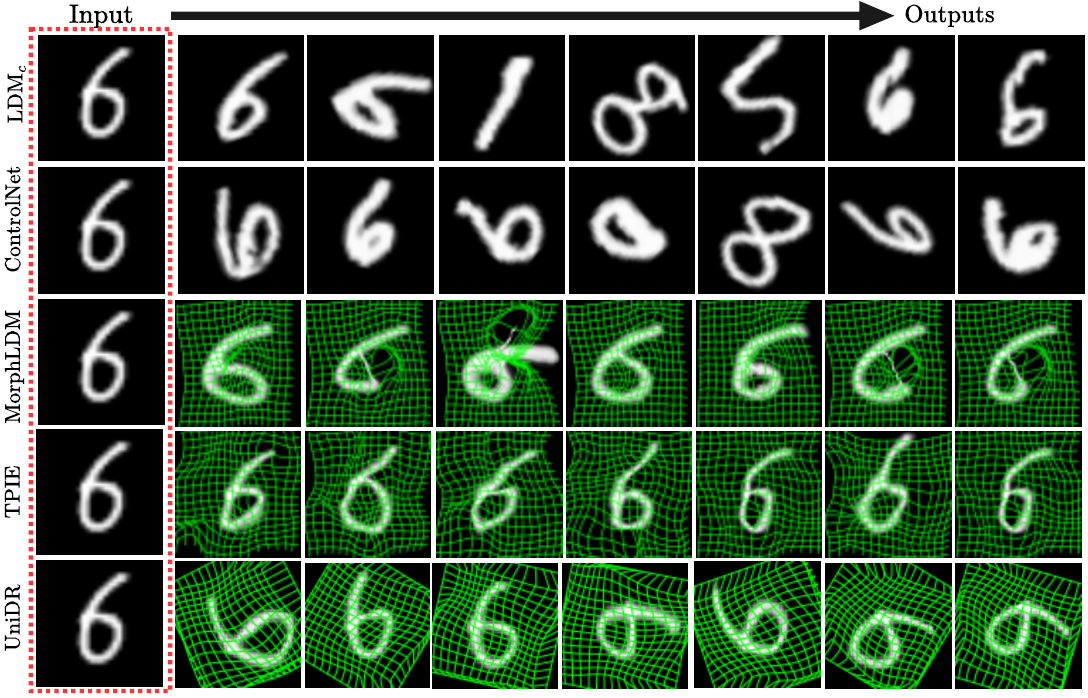}
\captionof{figure}{A comparison of generated samples on Roto-Translated MNIST. UniDR produces larger and more diverse rotations and translations while preserving digit topology.}
\label{fig:mnist}
\end{figure}

Tab.~\ref{tab:results_mnist} presents the quantitative evaluation of transformed images on synthetic roto-translated MNIST dataset. Note that the competing methods such as LDM \cite{rombach2022high} and ControlNet \cite{zhang2023adding} operate in intensity space without providing explicit deformation fields (marked as ``$-$'' for DetJac). Our results show that UniDR consistently achieves the best performance in all metrics, which reflects better image quality with preserved topology. 

Fig.~\ref{fig:mnist} visualizes exemplary generated samples from the MNIST dataset. Given the same input digit, the generative deformation baselines~\cite{wang2025generating,jayakumar2024tpie} generally preserve the topological structure better than the intensity-based diffusion baselines~\cite{rombach2022high,zhang2023adding}, which can introduce undesirable changes to the digit shape. Our UniDR preserves both the structural characteristics of input digits while generating diverse samples with explicitly learned affine transformations.
\begin{table*}[!htbp]
\centering
\setlength{\tabcolsep}{4pt}
\renewcommand{\arraystretch}{1.2}
\caption{\textbf{Quantitative evaluation on 3D Brain and 2D Cardiac MRI.} \textbf{Transf.} indicates whether a method produces an explicit deformation field. \textbf{Best} results are in bold and \underline{second-best} are underlined. The $^*$ indicates statistical significance ($p<0.05$).}
\label{tab:results_combined}
\resizebox{0.95\linewidth}{!}{%
\begin{tabular}{@{}l l c cccc cc c@{}}
\toprule
\multirow{2}{*}{\textbf{Dataset}} 
& \multirow{2}{*}{\textbf{Method}} 
& \multirow{2}{*}{\textbf{Transf.}}
& \multicolumn{4}{c}{\textbf{Image Fidelity}} 
& \multicolumn{2}{c}{\textbf{Image Diversity}} 
& \multicolumn{1}{c}{\textbf{Deformation Quality}} \\
\cmidrule(lr){4-7} 
\cmidrule(lr){8-9} 
\cmidrule(l){10-10}
& & 
& \textbf{FID}~\dn 
& \textbf{KID}~\dn 
& \textbf{PSNR}~\up 
& \textbf{SSIM}~\up
& \textbf{MS-SSIM}~\dn 
& \textbf{LPIPS}~\up
& \textbf{DetJac$_{-}$ (\%)}~\dn \\
\midrule
 
\multirow{6}{*}{\makecell[l]{3D Brain\\MRI}}
    & LDM~\cite{rombach2022high}     
  & \xmark 
  & \val{2.818}{0.163}  
  & \val{0.488}{0.03}  
  & \val{22.000}{2.810} 
  & \val{0.900}{0.029} 
  & \val{0.699}{0.22} 
  & \val{0.286}{0.17} 
  & \na \\
  
  & SADM~\cite{yoon2023sadm}     
  & \xmark 
  & \val{1.370}{0.550}  
  & \val{0.298}{0.050}  
  & \val{22.639}{4.313} 
  & \val{0.899}{0.051} 
  & \val{0.687}{0.10} 
  & \val{0.358}{0.10} 
  & \na \\

  & BrLP~\cite{puglisi2024enhancing}     
  & \xmark 
  & \val{1.093}{0.580}  
  & \val{0.083}{0.330}  
  & \val{23.151}{3.103} 
  & \val{0.913}{0.050}  
  & \val{0.691}{0.21} 
  & \val{0.308}{0.18} 
  & \na \\

  & MorphLDM~\cite{wang2025generating} 
  & \cmark 
  & \val{0.480}{0.125}  
  & \uval{0.077}{0.050} 
  & \val{25.091}{3.944} 
  & \val{0.933}{0.035}  
  & \val{0.687}{0.14} 
  & \val{0.355}{0.16} 
  & \val{0.0038}{0.0009} \\

  & TPIE~\cite{jayakumar2024tpie}   
  & \cmark 
  & \uval{0.433}{0.125} 
  & \val{0.078}{0.005}  
  & \uval{25.794}{3.877} 
  & \uval{0.937}{0.037} 
  & \uval{0.685}{0.10} 
  & \uval{0.375}{0.11} 
  & \uval{0.0013}{0.0001} \\

\rowcolor{ourgray}
  & \textbf{UniDR (Ours)}           
  & \cmark 
  & \bval{0.232}{0.089}$^*$ 
  & \bval{0.059}{0.045}$^*$ 
  & \bval{26.248}{3.970}$^*$ 
  & \bval{0.944}{0.037}$^*$ 
  & \bval{0.674}{0.22}$^*$ 
  & \bval{0.393}{0.13}$^*$ 
  & \bval{0.0009}{0.0003}$^*$ \\

\midrule
 
\multirow{6}{*}{\makecell[l]{2D Cardiac\\MRI}}
    & LDM~\cite{rombach2022high}     
  & \xmark 
  & \val{3.089}{0.254}  
  & \val{0.247}{0.0200}  
  & \val{17.260}{0.113} 
  & \val{0.511}{0.087} 
  & \val{0.797}{0.100} 
  & \val{0.220}{0.105} 
  & \na \\
  
  & ControlNet~\cite{zhang2023adding} 
  & \xmark 
  & \val{3.130}{0.780}  
  & \val{0.4870}{0.0300}  
  & \val{15.441}{1.032} 
  & \val{0.508}{0.068}  
  & \val{0.787}{0.110} 
  & \val{0.267}{0.160} 
  & \na \\

  & SADM~\cite{yoon2023sadm}       
  & \xmark 
  & \val{3.030}{0.321}  
  & \val{0.2270}{0.0150}  
  & \val{17.593}{1.732} 
  & \val{0.531}{0.065}  
  & \val{0.707}{0.010} 
  & \val{0.282}{0.100} 
  & \na \\

  & MorphLDM~\cite{wang2025generating} 
  & \cmark 
  & \val{0.689}{0.111}  
  & \val{0.0188}{0.0030}  
  & \uval{19.505}{1.889} 
  & \uval{0.554}{0.076} 
  & \uval{0.638}{0.070} 
  & \uval{0.342}{0.250} 
  & \val{0.0210}{0.010} \\

  & TPIE~\cite{jayakumar2024tpie}     
  & \cmark 
  & \uval{0.660}{0.287} 
  & \uval{0.0154}{0.0060} 
  & \val{17.909}{1.664} 
  & \val{0.537}{0.079}  
  & \val{0.644}{0.100} 
  & \val{0.338}{0.180} 
  & \uval{0.0003}{0.0001} \\

\rowcolor{ourgray}
  & \textbf{UniDR (Ours)}             
  & \cmark 
  & \bval{0.343}{0.120}$^*$ 
  & \bval{0.0112}{0.0010}$^*$ 
  & \bval{20.931}{1.958}$^*$ 
  & \bval{0.583}{0.084}$^*$ 
  & \bval{0.618}{0.100}$^*$ 
  & \bval{0.358}{0.210}$^*$ 
  & \bval{0.0001}{0.0002}$^*$ \\
\bottomrule
\end{tabular}%
}
\end{table*}

\begin{figure*}[!htbp]
   \centering
   \includegraphics[width=\linewidth, trim={0cm 0.0cm 0.0cm 0.0cm}, clip]{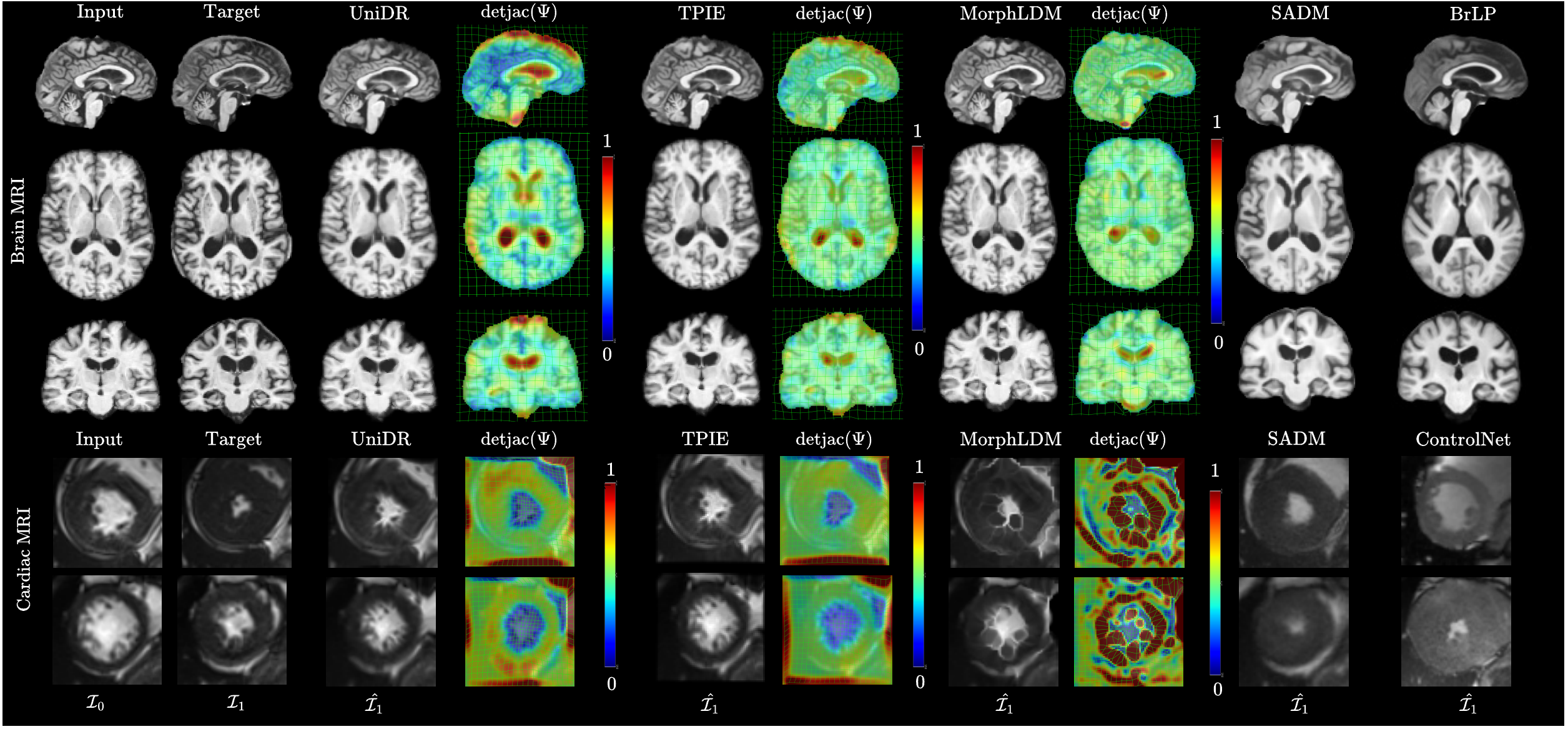}
   \captionof{figure}{Qualitative visualization of the generated samples. UniDR preserves the underlying topology with higher morphological details.}
   \label{fig:comparison_brain}
\end{figure*}

\noindent \textbf{Real Data Results.} Tab.~\ref{tab:results_combined} presents the quantitative evaluation of generated images on real-world brain MRIs and Cardiac MRI datasets. Compare with the two-step approaches, MorphLDM \cite{wang2025generating} and TPIE \cite{jayakumar2024tpie}, UniDR consistently achieves better performance in terms of image fidelity metrics and deformation quality. We use paired t-tests~\cite{mowery2011paired} comparing UniDR with each baseline. All comparisons yield $p < 0.05$ implying statistical significant. Similar comparison on brain MRI without pre-alignment are shown in Supplementary Sec.3.

Fig.~\ref{fig:comparison_brain} shows the generated brain and cardiac images from UniDR and the competing baselines, along with the DetJac map for generative deformation models. UniDR generates images that closely align with the corresponding target images, with smooth and spatially regular deformations reflected from the Jacobian map. More visualization results are included in Supplementary Sec.3.



Tab.~\ref{tab:dice_per_structure} reports the Dice scores between the segmentation masks propagated using the generated deformations and the corresponding ground-truth segmentation masks across anatomical structures in the brain and cardiac datasets. UniDR consistently achieves higher Dice scores than existing baselines, which indicates that its superior performance in synthesizing more anatomically faithful transformations. Additional dice score for the rest of the anatomical structures are provided in Supplementary Sec.~3. 

The top panel of Fig.~\ref{fig:mesh} displays representative comparisons between the ground-truth target anatomy and the propagated lateral ventricle obtained using UniDR and the second-best baseline. The bottom panel visualizes the signed local Euclidean surface distance~\cite{styner2006framework}, where positive values indicate that the propagated surface lies outside the ground-truth surface, while negative values indicate that it lies inside. It shows that the propagated lateral ventricle produced by UniDR closely aligns with the ground-truth surface and exhibits smaller and more spatially consistent surface deviations than the baseline.

Fig.~\ref{fig:anat_dice} demonstrates the applicability of the generated transformations for downstream segmentation task by showing the dice score for individual anatomical structure under two network backbones: UNet~\cite{ronneberger2015u} and Swin-UNETR~\cite{hatamizadeh2021swin}. The results demonstrate that training segmentation models with ground truth images augmented by UniDR generated samples (GT+Aug) consistently improves segemntation accuracy by $1\%$–$5\%$ across the evaluated brain structures (Fig.~\ref{fig:anat_dice}, Left). In a similar setting, on the ACDC cardiac dataset (Fig.~\ref{fig:anat_dice}, Right) the generated samples (GT+Aug) improves Dice by approximately $4.1\%$ for UNet and $5.6\%$ for Swin-UNETR.
\begin{table}[!t]
\centering
\setlength{\tabcolsep}{7pt}
\renewcommand{\arraystretch}{1.2}
\caption{Dice scores evaluated between the propagated masks using generative deformation models and ground-truth target. The $^*$ indicates statistical significance ($p<0.05$).}
\label{tab:dice_per_structure}
\resizebox{\linewidth}{!}{%
\begin{tabular}{@{}l cccc c@{}}
\toprule
\multirow{3}{*}{\textbf{Method}}
& \multicolumn{4}{c}{\textbf{Brain}} & \multicolumn{1}{c}{\textbf{Cardiac}} \\
\cmidrule(lr){2-5} \cmidrule(l){6-6}
& \makecell{Lateral\\Ventricles} & Hippocampus & Thalamus & Cortex & \makecell{Left\\Ventricle} \\
\midrule
MorphLDM~\cite{wang2025generating}
  & $0.78 \pm \scriptsize{0.15}$ & $0.77 \pm \scriptsize{0.07}$ & $0.80 \pm \scriptsize{0.07}$
  & $0.78 \pm \scriptsize{0.07}$ & $0.74 \pm \scriptsize{0.08}$ \\
TPIE~\cite{jayakumar2024tpie}
  & $0.82 \pm \scriptsize{0.11}$ & $\mathbf{0.79 \pm \scriptsize{0.07}}$ & $0.83 \pm \scriptsize{0.06}$
  & $0.80 \pm \scriptsize{0.07}$ & $0.73 \pm \scriptsize{0.07}$ \\
  \rowcolor{ourgray}
\textbf{UniDR (Ours)}
  & $\mathbf{0.84 \pm \scriptsize{0.09}}^*$ & $0.78 \pm \scriptsize{0.07}$ & $\mathbf{0.84 \pm \scriptsize{0.06}}^*$
  & $\mathbf{0.81 \pm \scriptsize{0.07}}$ & $\mathbf{0.76 \pm \scriptsize{0.08}}^*$ \\
\bottomrule
\end{tabular}}
\end{table}

\begin{figure}[!t]
    \centering
    \includegraphics[width=\linewidth, trim={0.0cm 0.0cm 0.0cm 0.0cm}, clip]{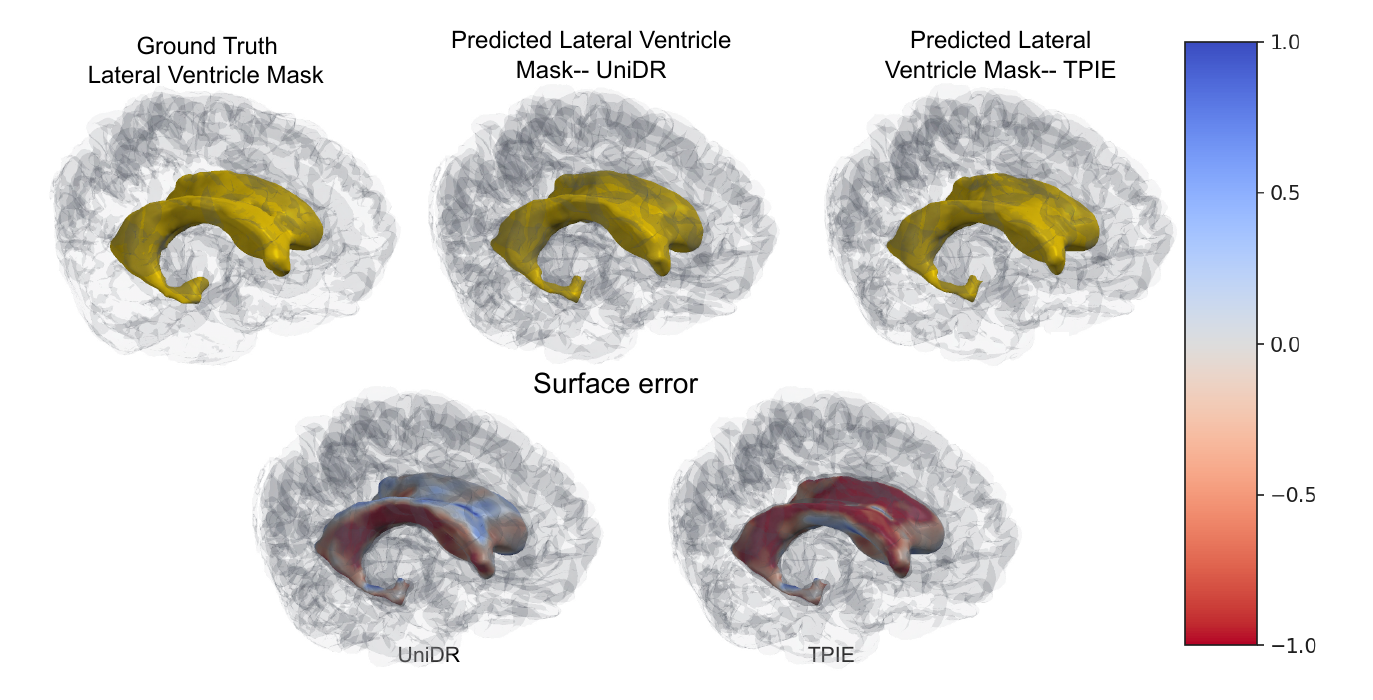}
    \caption{Visualization of the transformed lateral ventricle shapes. The learned transformations from UniDR and TPIE are applied to the input lateral ventricle shape and compared with the ground truth shape. The colorbar denotes the propagated surface error.}
    \label{fig:mesh}
\end{figure}

\begin{figure}[!t]
    \centering
    \includegraphics[width=\linewidth, trim={0.0cm 0.0cm 0.0cm 0.0cm}, clip]{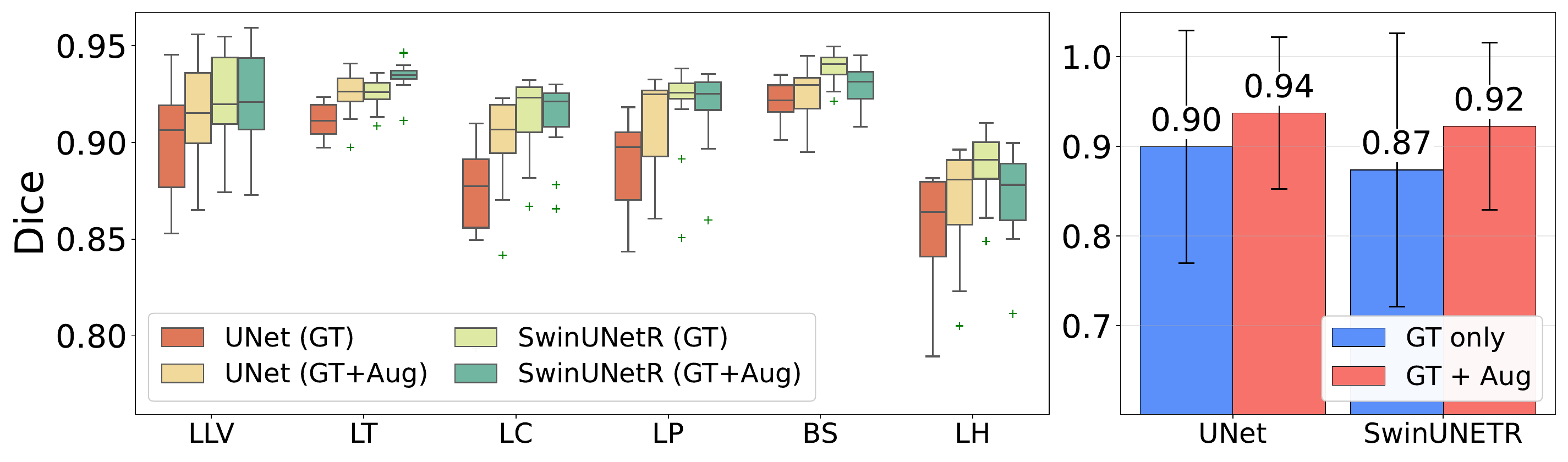}
    \caption{Downstream segmentation performance of models trained with ground-truth (GT) images and GT images augmented with UniDR-generated samples (GT+UniDR), reported as Dice scores for individual brain structures (Left) and LV myocardium segmentation (Right).}
    \label{fig:anat_dice}
\end{figure}

\begin{figure}[!t]
    \centering
    \includegraphics[width=\linewidth, trim={0.0cm 0.0cm 0.0cm 0.0cm}, clip]{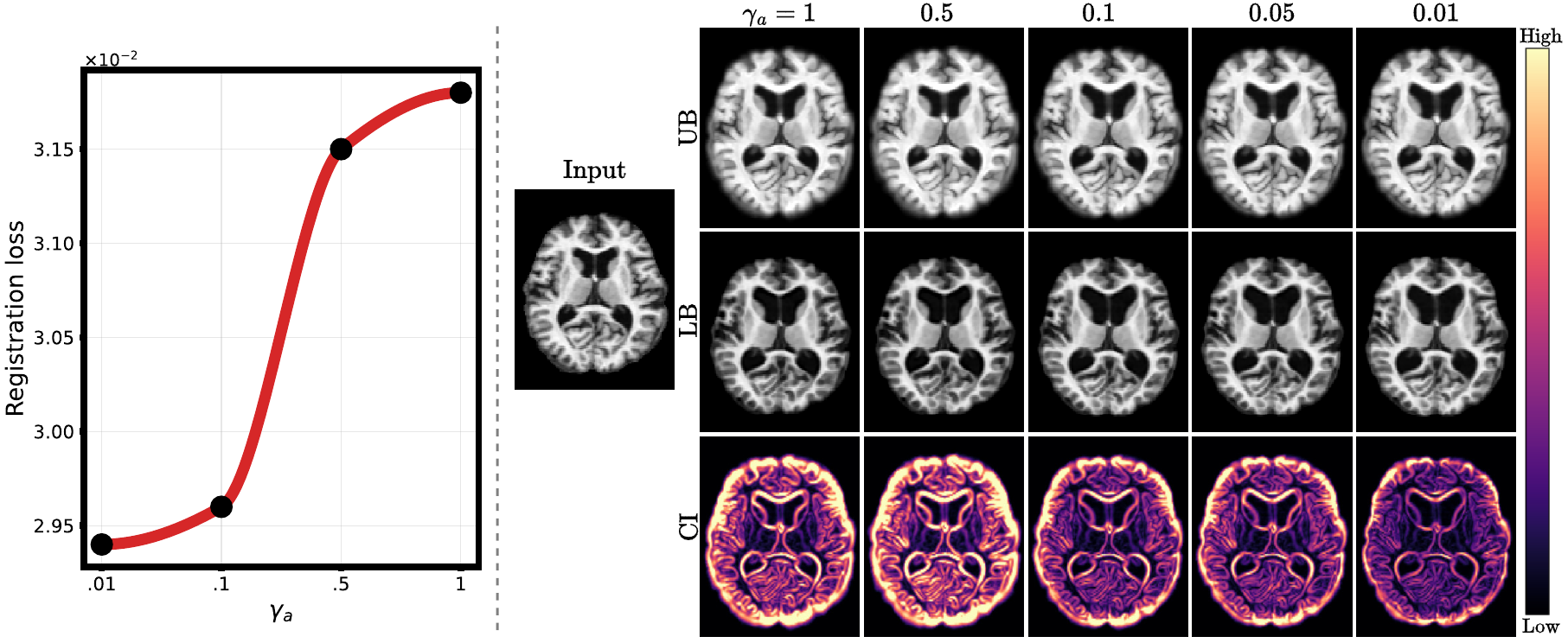}
    \caption{Visualization of confidence map for a given input image. We show the upper (UB), lower (LB) bound and confidence interval (CI) with 95\% level based on $1000$ generated samples.} 
    \label{fig:ablation}    
\end{figure}

\noindent \textbf{Ablation on the weighting parameter between registration and diffusion.} Fig.~\ref{fig:ablation} investigates the effect of the weighting parameter $\gamma_a$ that control the effect of the registration term and the score matching objective. According to Eq.~\eqref{eq:loss}, increasing $\gamma_a$ reduces the influence of the registration objective, hence increasing the registration loss (see Fig.~\ref{fig:ablation}, Left). Consequently, our model generates diverse transformations with substantial variability across samples as seen from the confidence map (Fig.~\ref{fig:ablation}, Right).

\begin{table}[!t]
\centering
\small
\setlength{\tabcolsep}{5pt}
\renewcommand{\arraystretch}{1.1}
\caption{Ablation of two-step approach vs joint optimization.}
\label{tab:fixed_v0_ablation}

\resizebox{\columnwidth}{!}{%
\begin{tabular}{lccrrr}
\toprule
Method 
& Fixed $\Gamma^i_0$
& Backbone
& FID $\downarrow$
& PSNR $\uparrow$
& DSC $\uparrow$ \\
\midrule

UniDR$_{TM}$
& $\checkmark$
& TransMorph~\cite{chen2022transmorph}
& $0.397$ & $25.87$ & $0.785$ \\

UniDR$_{uni}$
& $\checkmark$
& uniGradICON~\cite{tian2024unigradicon}
& $0.590$ & $25.78$ & $0.780$ \\

UniDR$_{R2N}$
& $\checkmark$
& R2Net~\cite{joshi2022diffeomorphic}
& $0.441$ & $25.28$ & $0.773$ \\

UniDR$_{VFA}$
& $\checkmark$
& VFA~\cite{liu2024vector}
& $0.450$ & $25.23$ & $0.751$ \\

\midrule

UniDR
& $\times$
& \textbf{$\times$}
& $\mathbf{0.232}$
& $\mathbf{26.248}$
& $\mathbf{0.799}$ \\

\bottomrule
\end{tabular}%
}
\end{table}

\noindent \textbf{Ablation on joint optimization vs. two-step approach.} Tab.~\ref{tab:fixed_v0_ablation} presents an ablation study of UniDR with two-step variants using different pretrained registration backbones. Specifically, we first estimate the deformation using TransMorph~\cite{chen2022transmorph}, R2Net~\cite{joshi2022diffeomorphic}, uniGradICON~\cite{tian2024unigradicon}, and VFA~\cite{liu2024vector}, and then use these estimated transformations as supervision to train UniDR by optimizing only the score-matching loss. Since uniGradICON is a pretrained foundation model, we use it directly in an off-the-shelf manner, whereas the other registration backbones are separately trained on the ADNI dataset. Compared with all variants relying on fixed supervision from pretrained registration models, UniDR consistently achieves better performance demonstrating the advantage of joint optimization.



\section{Conclusion \& Limitation Discussion}
In this work, we introduced UniDR, a novel diffusion model that learns the distribution of transformations by unifying diffusion with a registration objective, without relying on a separate registration backbone. This enables UniDR to learn and progressively refine underlying deformation through a joint objective. Experiments demonstrate its effectiveness across various medical applications, while the generated transformations can also be directly leveraged for downstream tasks such as image segmentation.

Our UniDR currently focuses on generative modeling of transformations, while recognizing that image intensity variations capturing appearance changes are also important for modeling heterogeneous imaging data. A promising direction for future work is to jointly model geometric deformations and appearance variations, enabling more comprehensive and flexible generation and potentially improving the adaptability and generalizability of UniDR across heterogeneous imaging domains.

\section{Acknowledgements}
This work was supported by the funding from the NSF CAREER 2239977, NIH R01HL181107, and NSF-SCH 2501059.

{
    \small
    \bibliographystyle{ieeenat_fullname}
    \bibliography{main}
}

\clearpage

\appendix

\section{Experimental Setup}
\noindent \textbf{Evaluation of the fidelity of transformed images.} We evaluate the quality of our learned transformations on all datasets by quantifying Fréchet Inception Distance (FID) \cite{heusel2017gans}, Kernel Inception Distance (KID) \cite{DBLP:conf/iclr/BinkowskiSAG18},  Multi-scale Structural Similarity Index Measure (MS-SSIM) \cite{friedrich2024wdm,wang2025generating}, and Learned Perceptual Image Patch Similarity (LPIPS) \cite{zhang2018unreasonable} for all baseline models. 

\begin{itemize}
    \item \textbf{Fréchet Inception Distance (FID).} The FID quantifies image fidelity by measuring the similarity between the distributions of real and generated images. Following common practice in 3D medical imaging~\cite{dorjsembe2024conditional,sun2022hierarchical}, we extract high-level features from real and transformed images using an intermediate activation layer of a pretrained Med3D network~\cite{chen2019med3d}. We then fit multivariate Gaussian distributions to the extracted feature representations and compute the FID between these distributions. 
    \item \textbf{Kernel Inception Distance (KID).} We compute the KID score as the squared maximum mean discrepancy (MMD) between the Inception features of real and generated images. This provides an unbiased estimate of the distribution distance and is particularly robust for smaller sample sizes.
    \item \textbf{Peak Signal-to-Noise Ratio (PSNR).} We compute PSNR between each transformed image and its corresponding target image to quantify the fidelity of the pixel-level reconstruction.
    \item \textbf{Structural Similarity Index Measure (SSIM).} We compute SSIM between each transformed image and its corresponding target image to assess structural similarity in terms of local luminance, contrast, and image structure. 
    \item \textbf{Multi-Scale Structural Similarity (MS-SSIM).} We measure the MS-SSIM score to evaluate the diversity of the transformed images~\cite{friedrich2024wdm,wang2025generating} by computing pairwise comparisons between all generated images.
    \item \textbf{Learned Perceptual Image Patch Similarity (LPIPS).} We employ a pretrained AlexNet to quantify perceptual diversity in the feature space.
\end{itemize}

We also evaluate the smoothness of the sampled deformation by examining its determinant of the Jacobian (DetJac), which provides a direct assessment of whether the generated transformations preserve anatomical topology or introduce undesirable artifacts. In particular, A DetJac value of 1 indicates local volume preservation, values less than 1 correspond to local contraction, and values greater than 1 indicate local expansion. In contrast, negative values indicate a violation of anatomical topology. \\

\noindent \textbf{Implementation details.} We set the weighting parameters $\gamma_a$ value as $1$ and $0.1$ for the Brain and Cardiac dataset respectively. The $\gamma_w$ is set as $0.5$ and $0.05$ for Brain and Cardiac dataset, respectively. For the Roto-Translated MNIST experiment, we set $\gamma_a$= 1 and $\gamma_w$= 0.5. For the ADNI brain data set, we use age at the two longitudinal time points, gender, and disease label as metadata, and for cardiac data set, we use the diagnosis label as metadata. 

Our model is optimized using the Adam optimizer with a learning rate of $10^{-4}$, a batch size of $4$, and is trained for $100$K training steps. We use a 3D UNet-based denoiser following the OpenAI/guided-diffusion repository for 3D brain MRI experiments~\cite{dhariwal2021diffusion}. For the forward process, we adopt the sigmoid-beta noise scheduler similar to~\cite{jabri2023scalable}, and for sampling, we use $1000$ denoising steps as described in~\cite{song2020denoising}. 

We use $70/30$ train/test subject-wise split in all our experiments. All models are implemented in PyTorch and trained using a single NVIDIA A100 GPU. All reported performance is averaged over $10$ independent sampling runs from the same model to ensure statistical reliability. 

\section{Downstream Segmentation Result}
Fig.~\ref{fig:anat_dice} shows the details dice score across all the anatomical structure for the downstream brain segemntation task. This indicate that UniDR captures meaningful anatomical variability and hemispheric symmetry to improve generalization.

\begin{figure}[h]
    \centering
    \includegraphics[width=\linewidth, trim={0.0cm 0.0cm 0.0cm 0.0cm}, clip]{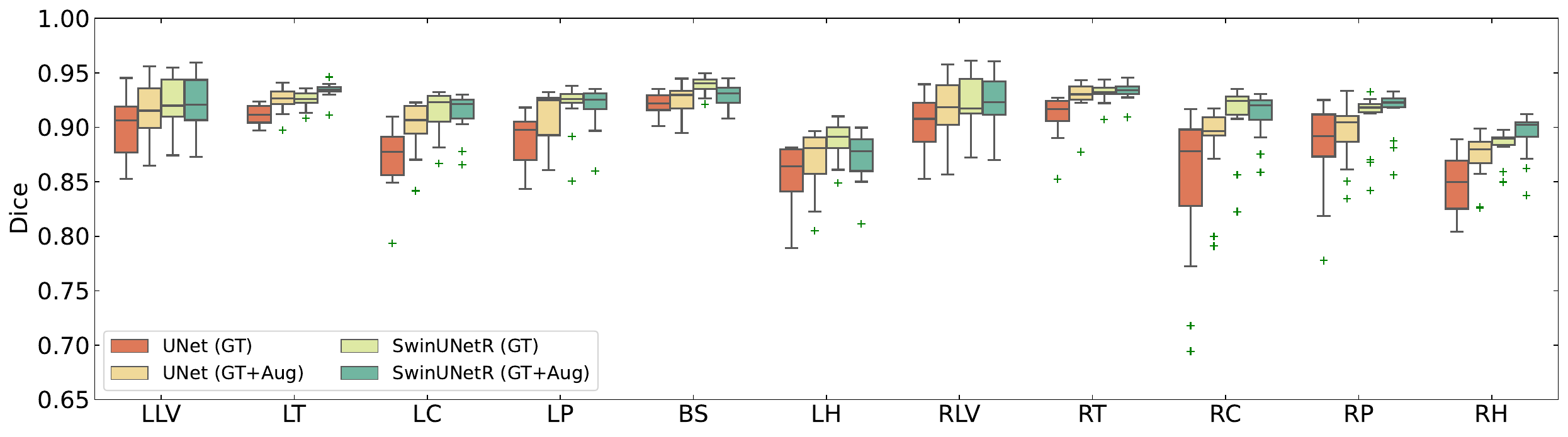}
    \caption{Dice score comparison across key brain structures for UNet and SwinUNETR, trained on ground-truth alone (GT) and with our generated images (GT+Aug). UniDR-augmented training yields consistent gains, most notably in complex structures like the hippocampus and thalamus—while preserving anatomical consistency across hemispheric structures (LWM/RWM, LLV/RLV, LT/RT, LC/RC, LP/RP, LH/RH) and the brain stem (BS).}
    \label{fig:anat_dice}
\end{figure} 



\section{Additional Results on ADNI Brain}
\begin{figure}[!t]
   \centering
   \includegraphics[width=\linewidth, trim={0.0cm 0.0cm 0.0cm 0.0cm}, clip]{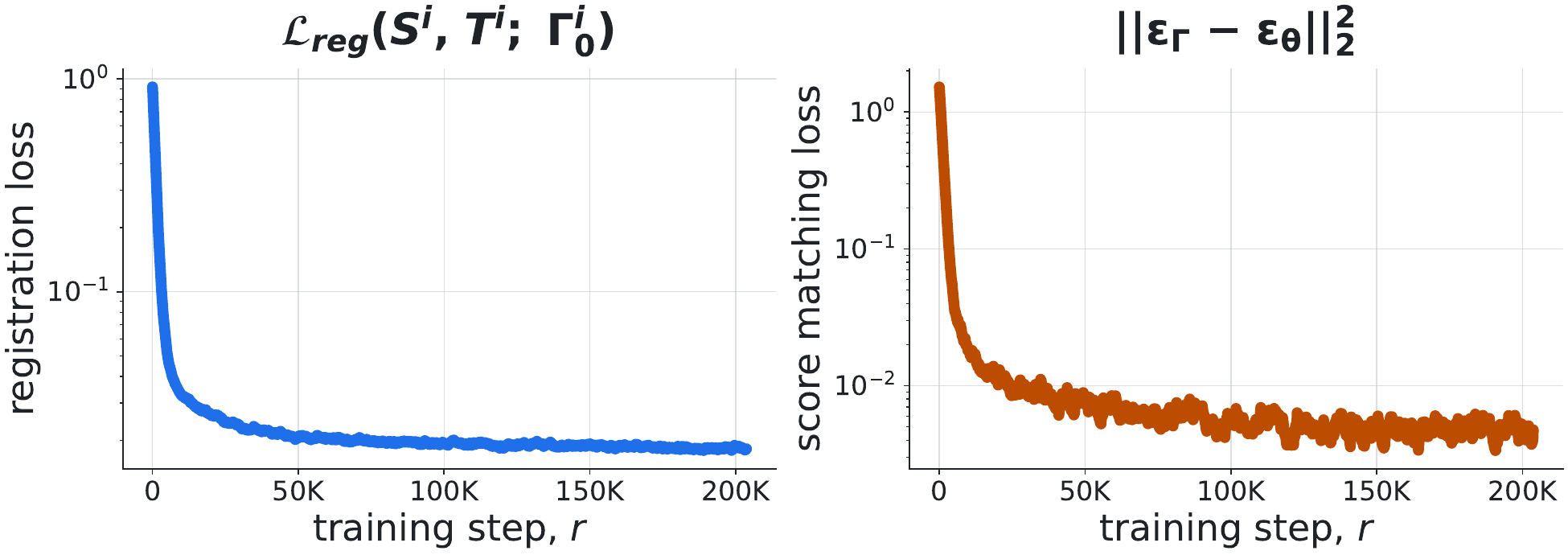}
   \caption{Training dynamics of the joint objective on ADNI brain dataset. Registration loss of the predicted clean transformation $\hat{\Gamma}^i_0$ (Left) and score-matching loss (right) versus the traning step r. Both loss gradually decreases and plateau, indicating the joint objective is gradually minimized.}
   \label{fig:loss_curve}
\end{figure}

In Fig.~\ref{fig:loss_curve}, we show the training loss curve of the reigstration loss and the score-matching loss as traininig step 
progresses. As it can be seen, both losses gradually decreases and plateau as the training progresses.

Tab.~\ref{tab:dice} provides the dice score for each anatomical structure for the ADNI brain data set. We calculate the dice score between the ground-truth segmentation masks and the propagated segmentation mask of the input image using the synthesized transformation. The ground truth segmentation mask is extracted using the FreeSurfer software. The synthesized transformation is directly applied to the corresponding input segmentation mask, and the Dice score is then computed between the transformed mask and the target ground-truth segmentation. UniDR consistently achieves higher Dice scores than the existing deformation-based baselines across most anatomical structures, indicating the learned transformations better capture the anatomical structure of the target mask.

\begin{figure*}[!htbp]
   \centering
   \includegraphics[width=0.9\linewidth, trim={0cm 0.0cm 0.0cm 0.0cm}, clip]{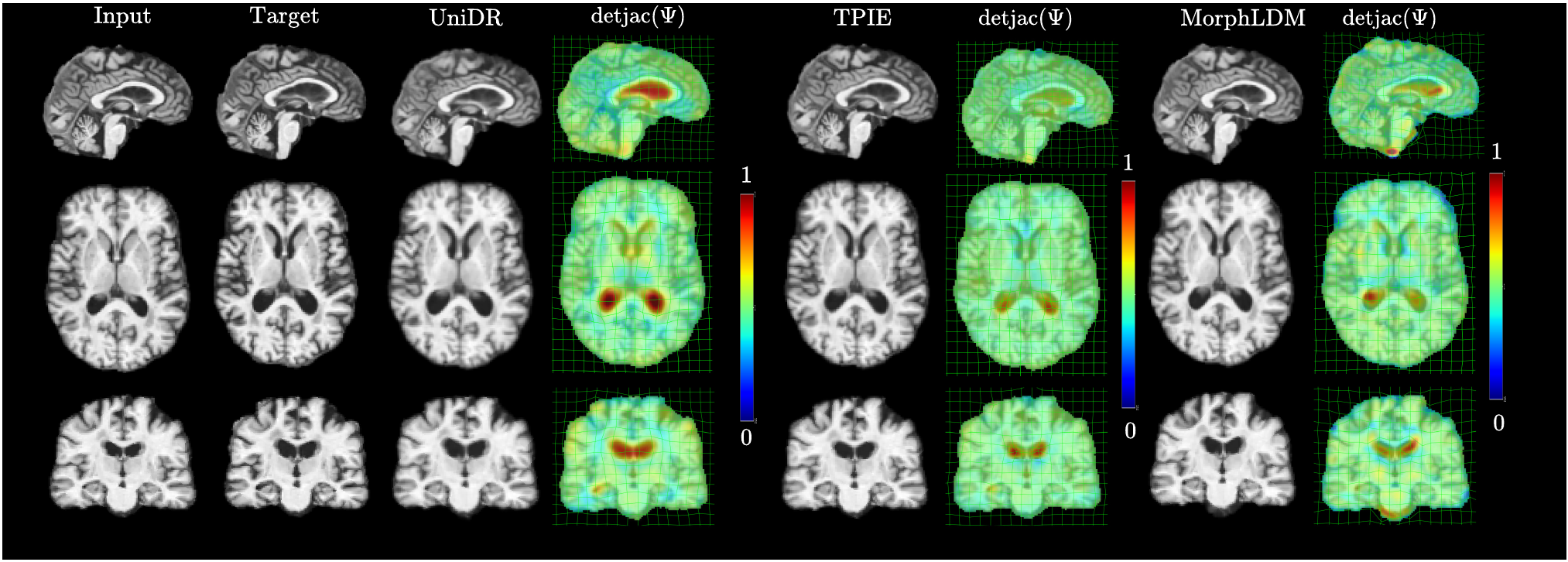}
   \includegraphics[width=0.9\linewidth, trim={0cm 0.0cm 0.0cm 0.0cm}, clip]{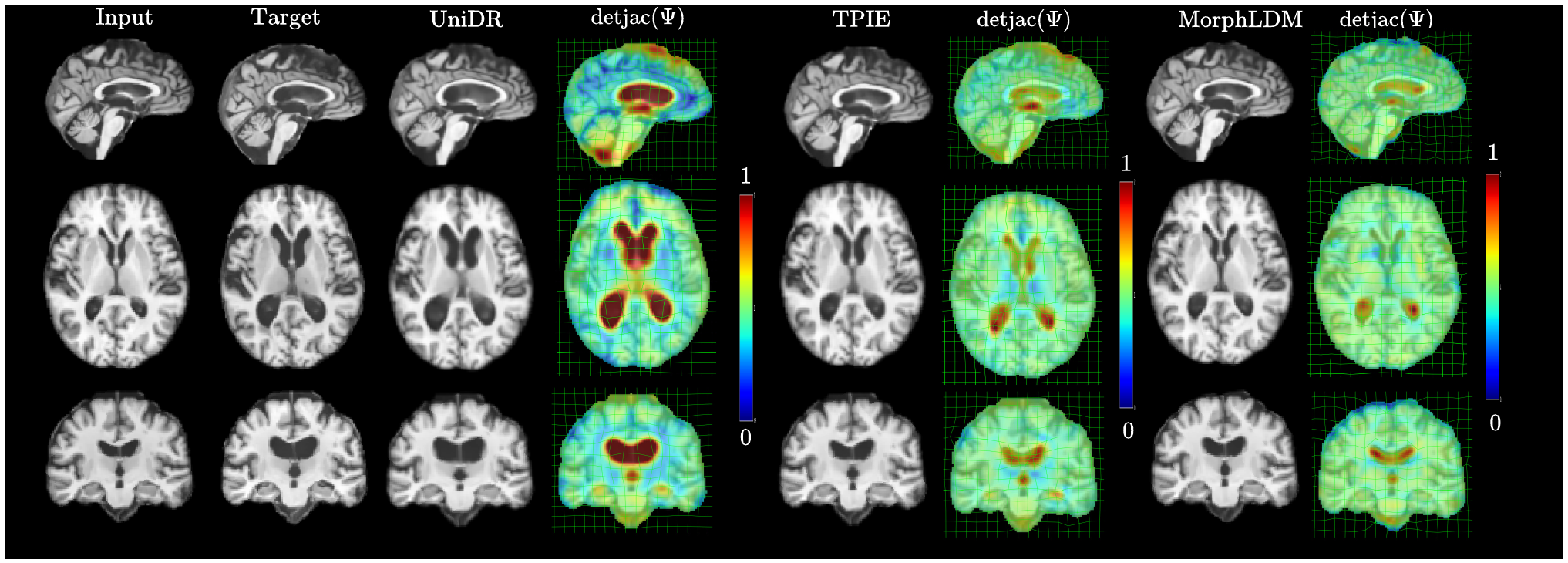}
   \includegraphics[width=0.9\linewidth, trim={0cm 0.0cm 0.0cm 0.0cm}, clip]{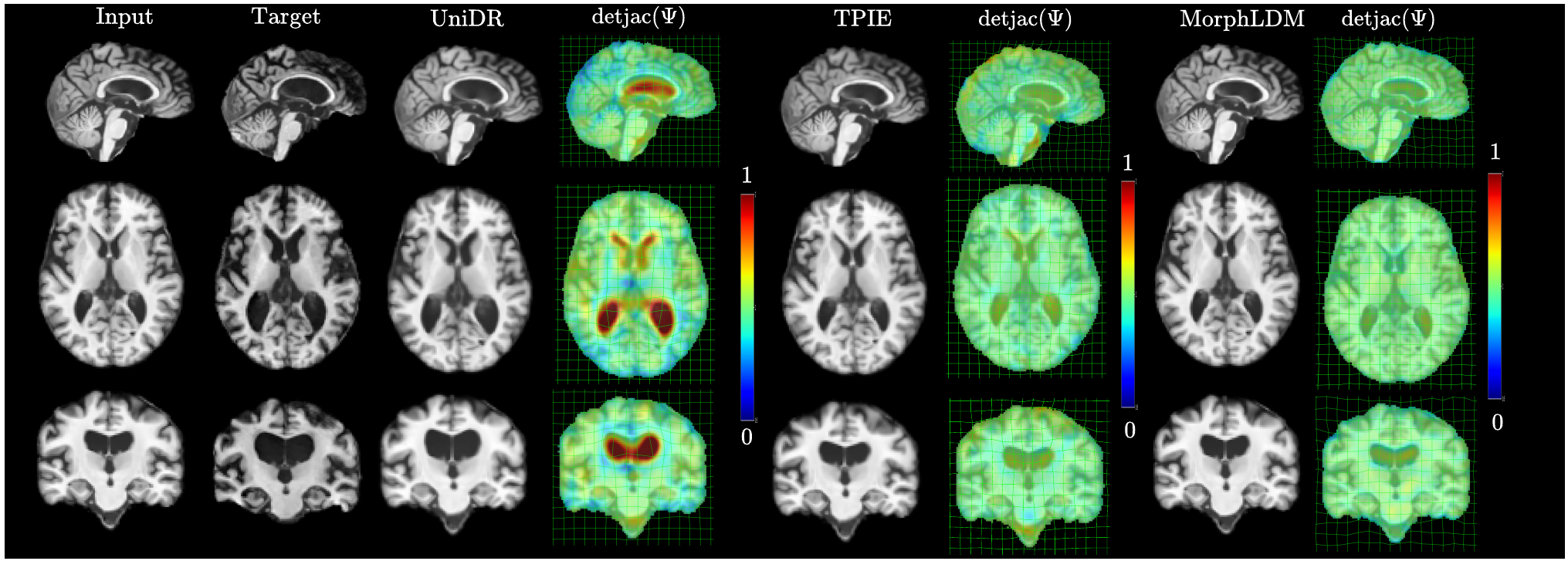}
   \includegraphics[width=0.9\linewidth, trim={0cm 0.0cm 0.0cm 0.0cm}, clip]{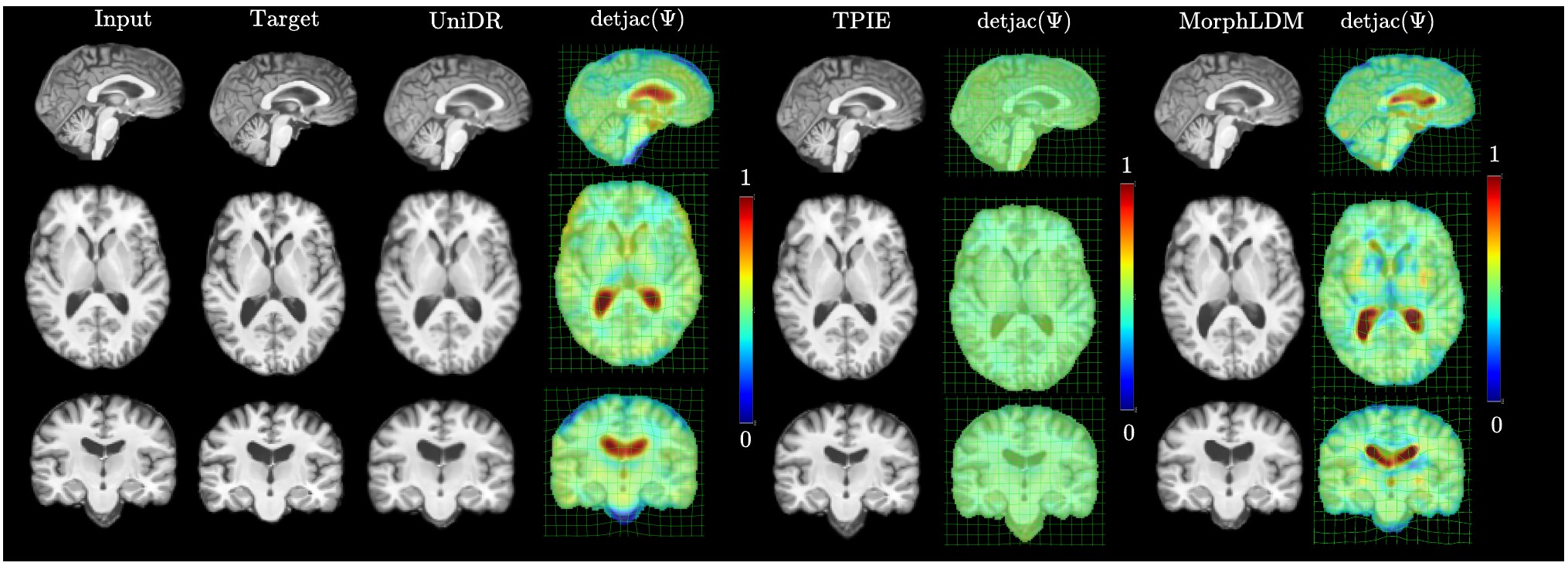}
   \captionof{figure}{Qualitative visualization of the generated samples for the transformation based generative models.}
   \label{fig:comparison_brain_supp}
\end{figure*}

Fig.~\ref{fig:comparison_brain_supp} provides qualitative comparisons of the generated samples from transformation-based generative models, including TPIE~\cite{jayakumar2024tpie} and MorphLDM~\cite{wang2025generating}. The results show that UniDR generates more anatomically plausible deformations that better capture the structural differences between the source and target images.

Tab.~\ref{tab:results_unaligned} reports results on unaligned ADNI brain MRI data. In contrast to our model UniDR, the two other baselines lack explicit affine transformation components and must therefore capture global mis-alignments entirely through their deformation fields. By jointly modeling affine and deformable transformations, UniDR consistently achieves superior image fidelity and deformation quality.

\begin{figure*}[!htbp]
   \centering
   \includegraphics[width=0.95\linewidth, trim={0cm 0.0cm 0.0cm 0.0cm}, clip]{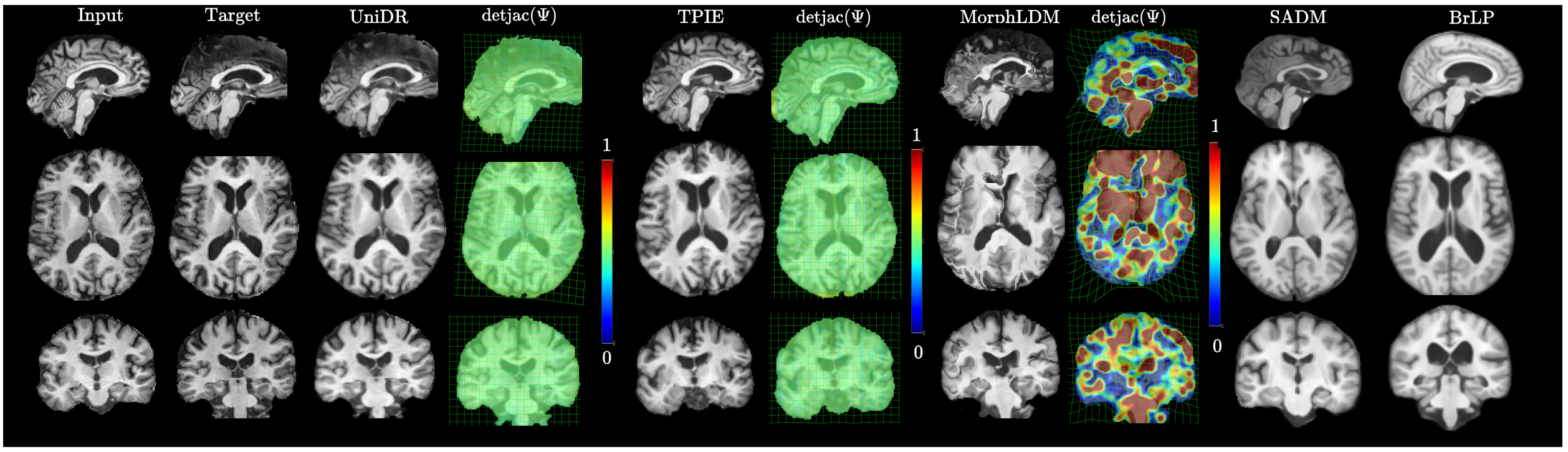}
   \captionof{figure}{Qualitative visualization of the generated samples on unaligned ADNI brain MRI. UniDR preserves the underlying topology with higher morphological details.}
   \label{fig:comparison_brain_unaligned}
\end{figure*}

Fig.~\ref{fig:comparison_brain_unaligned} visualizes a comparison of the generated images from UniDR and baseline methods, including both transformation-based and image-intensity-based generative models, on unaligned brain MRI data.

\begin{table}[t]
\centering
\caption{Per-structure Dice scores (higher is better, $\uparrow$) on the synthesized images, reported as mean $\pm$ std across subjects. \textbf{Bold} indicates the best result per structure.}
\label{tab:dice}
\setlength{\tabcolsep}{5pt}
\resizebox{\columnwidth}{!}{%
\begin{tabular}{l ccc}
\toprule
Structure & MorphLDM & TPIE & UniDR (Ours) \\
\midrule
Lateral Ventricles & $0.7884$ {\scriptsize$\pm 0.116$} & $0.8229$ {\scriptsize$\pm 0.117$} & $\mathbf{0.8497}$ {\scriptsize$\pm 0.099$} \\
Hippocampus        & $0.7683$ {\scriptsize$\pm 0.073$} & $\mathbf{0.7900}$ {\scriptsize$\pm 0.078$} & $0.7886$ {\scriptsize$\pm 0.078$} \\
Cortex             & $0.7846$ {\scriptsize$\pm 0.078$} & $0.8051$ {\scriptsize$\pm 0.075$} & $\mathbf{0.8119}$ {\scriptsize$\pm 0.076$} \\
Cerebral WM        & $0.7936$ {\scriptsize$\pm 0.078$} & $0.8126$ {\scriptsize$\pm 0.079$} & $\mathbf{0.8250}$ {\scriptsize$\pm 0.078$} \\
Amygdala           & $0.7328$ {\scriptsize$\pm 0.090$} & $0.7380$ {\scriptsize$\pm 0.096$} & $\mathbf{0.7401}$ {\scriptsize$\pm 0.093$} \\
Thalamus           & $0.8083$ {\scriptsize$\pm 0.070$} & $0.8388$ {\scriptsize$\pm 0.065$} & $\mathbf{0.8415}$ {\scriptsize$\pm 0.065$} \\
Caudate            & $0.5854$ {\scriptsize$\pm 0.179$} & $0.6548$ {\scriptsize$\pm 0.168$} & $\mathbf{0.6897}$ {\scriptsize$\pm 0.156$} \\
Putamen            & $0.7928$ {\scriptsize$\pm 0.083$} & $\mathbf{0.8015}$ {\scriptsize$\pm 0.091$} & $0.7997$ {\scriptsize$\pm 0.099$} \\
\midrule
Average            & $0.7568$ & $0.7830$ & $\mathbf{0.7933}$ \\
\bottomrule
\end{tabular}}
\end{table}

\begin{table*}[!htbp]
\centering
\setlength{\tabcolsep}{4pt}
\renewcommand{\arraystretch}{1.2}
\caption{\textbf{Quantitative evaluation on unaligned 3D ADNI Brain MRI.} \textbf{Transf.} indicates whether a method produces an explicit deformation field. \textbf{Best} results are in bold and \underline{second-best} are underlined. $^*$ indicates statistical significance ($p<0.05$).}
\label{tab:results_unaligned}
\resizebox{0.65\linewidth}{!}{%
\begin{tabular}{@{}l c ccc c@{}}
\toprule
\textbf{Method} 
& \textbf{Transf.}
& \textbf{FID}~\dn 
& \textbf{PSNR}~\up 
& \textbf{SSIM}~\up
& \textbf{DetJac$_{-}$ (\%)}~\dn \\
\midrule

  
SADM~\cite{yoon2023sadm}     
  & \xmark 
  & \val{3.510}{0.280}  
  & \val{18.200}{1.640} 
  & \val{0.694}{0.036} 
  & \na \\

BrLP~\cite{puglisi2024enhancing}     
  & \xmark 
  & \val{3.685}{0.220}  
  & \val{16.667}{1.496} 
  & \val{0.703}{0.043}  
  & \na \\

MorphLDM~\cite{wang2025generating} 
  & \cmark 
  & \uval{0.550}{0.105}  
  & \uval{19.487}{2.632} 
  & \uval{0.706}{0.063}  
  & \uval{0.503}{0.1268} \\

TPIE~\cite{jayakumar2024tpie}   
  & \cmark 
  & \val{0.577}{0.080} 
  & \val{18.404}{1.660} 
  & \val{0.687}{0.052} 
  & \val{0.3730}{0.3240} \\

\rowcolor{ourgray}
\textbf{UniDR (Ours)}           
  & \cmark 
  & \bval{0.301}{0.059}$^*$ 
  & \bval{20.163}{3.136}$^*$ 
  & \bval{0.749}{0.104}$^*$ 
  & \bval{0.0008}{0.0004}$^*$ \\
  
\bottomrule
\end{tabular}%
}
\end{table*}

\section{Implementation Details of the Registration Backbones}
This section provides the hyperparameter setup for the registration backbones in Tab. 4 (main paper). All models are validated on $128^3$ image volume with intensity normalization. The first time frame is used as the source and the subsequent longitudinal time frames are used as the target. 

\paragraph{TransMorph~\cite{chen2022transmorph}.} We train a 3D TransMorph Swin-transformer that uses a $2$-channel input of size $128^3$, patch size $4$, embedding dimension $48$, stage depths $(2,2,4,2)$, attention heads $(4,4,8,8)$, and window size $(4,4,4)$. The MLP ratio and patch-merging reduction factor are both $4$. Relative position encoding and patch normalization are enabled, while absolute and sinusoidal encodings are disabled. We use stochastic depth $0.3$, a registration-head width of $16$, and both transformer and convolutional skip connections. \textbf{Training:} We use Adam with a learning rate of $10^{-4}$ and no weight decay for $300$ epochs. The objective combines MSE similarity loss (weight $1.0$) and diffusion regularization (weight $0.01$), with $7$ integration steps to obtain the final transformation.

\paragraph{VFA~\cite{liu2024vector}.}
We initialize Vector Field Attention (VFA) using the released LUMIR brain-MRI weights (v0.0.6) and fine-tune it on the ADNI dataset. The architecture uses $4$ downsampling levels, $8$ starting channels, and $8$ matching channels, with an initialization scale of $0.1$. It accepts single-channel images and uses no skip connections, affine stage, or integration steps. We fine-tune the model for $1000$ epochs with a batch size of $1$, a learning rate of $10^{-4}$, and half-precision (fp16) computation. The training objective combines normalized cross-correlation (NCC), computed with a window size of $9$, and a gradient penalty, each weighted by $1.0$. The network is fully convolutional and dynamically constructs coordinate grids during execution.

\paragraph{R2Net~\cite{joshi2022diffeomorphic}.} R2Net releases only Keras/TF1-style code. We reimplement it in PyTorch following the authors' network structure: a VoxelMorph-2 U-Net over concat(source, target) with encoder widths $(16,32,32,32)$ and decoder widths $(32,32,32,32,16,16,3)$, a bias-free $3\times3\times3$ head giving the initial field scaled by $1/2$, followed by seven Lipschitz-continuous residual blocks (spectrally-normalised $3^3$ conv $\rightarrow$ LeakyReLU$(0.2)$ $\rightarrow$ spectrally-normalised $1^3$ conv $\rightarrow$ $\tanh$ $\rightarrow$ $1/2$), whose composition is diffeomorphic without scaling-and-squaring. Trained on the same $1815$ ADNI pairs: Adam, learning rate $10^{-4}$, batch size $1$, $150$ epochs, gradient-norm clipping at $1.0$, MSE similarity loss with $\sigma=0.02$, a folding penalty (weight $1.0$) and a gradient penalty (weight $1.0$).




\end{document}